%% file: main.tex
\documentclass{article}
\usepackage{spconf,amsmath,graphicx}
\usepackage{times}
\usepackage{soul}
\usepackage{url}
\usepackage[hidelinks]{hyperref}
\usepackage[utf8]{inputenc}
\usepackage[small]{caption}
\usepackage{amsthm}
\usepackage{booktabs}
\usepackage{algorithm}
\usepackage{algorithmic}
\usepackage{epsfig}
\usepackage{amssymb}
\usepackage{multirow}
\usepackage{xcolor}
\usepackage{adjustbox}

\title{Domain Adaptation for Learning Generator from Paired Few-Shot Data}
\name{Chun-Chih Teng$^{\dagger}$ \qquad Pin-Yu Chen$^{\ddagger}$ \qquad Wei-Chen Chiu$^{\dagger}$}
\address{$^{\dagger}$National Chiao Tung University\thanks{National Chiao Tung University and National Yang-Ming University are merged into National Yang Ming Chiao Tung University in February 2021.} \qquad $^{\ddagger}$IBM Research}  

\begin{document}
\maketitle
\begin{abstract}
We propose a Paired Few-shot GAN (PFS-GAN) model for learning generators with sufficient source data and a few target data. While generative model learning typically needs large-scale training data, our PFS-GAN not only uses the concept of few-shot learning but also domain shift to transfer the knowledge across domains, which alleviates the issue of obtaining low-quality generator when only trained with target domain data.
The cross-domain datasets are assumed to have two properties: (1) each target-domain sample has its source-domain correspondence and (2) two domains share similar content information but different appearance. Our PFS-GAN aims to learn the disentangled representation from images, which composed of domain-invariant content features and domain-specific appearance features. Furthermore, a relation loss is introduced on the content features while shifting the appearance features to increase the structural diversity. Extensive experiments show that our method has better quantitative and qualitative results on the generated target-domain data with higher diversity in comparison to several baselines.
\end{abstract}

\input{intro.tex}

\input{method.tex}

\input{exp.tex}

\input{conclusion.tex}

\bibliographystyle{IEEEbib}
\bibliography{egbib}

\end{document}

%% file: intro.tex
\section{Introduction}

Generative Adversarial Network (GAN)~\cite{goodfellow2014generative} is one of the most popular generative models based on deep learning nowadays, in which its learning process typically relies on sufficient amount of training data in order to generate realistic and diverse data samples. However, in some tasks such as medical applications, the training data are often scarce and the cost of acquiring new data samples is expensive, if not impossible. Consequently, the restriction of limited training data size may largely degrade the performance of GAN thus leading to mode collapse. To overcome this challenge, we propose to use domain adaptation \cite{vinyals2016matching,zhu2017unpaired,sung2018learning,xian2019f} to transfer the knowledge of generative process from another data collection with rich information (\textbf{\textit{source domain}}) to the target task (\textbf{\textit{target domain}}), where these two domains share some similarities but follow two different data distributions, also known as \textit{domain shift}. 
While domain adaptation has mostly been used for classification~\cite{motiian2017few,liu2019butterfly,tan2019weakly}, its application on learning generators with limited data is still challenging and not well-explored.
It is also worth noting that: Image-to-image translation (e.g., CycleGAN~\cite{isola2017image}) enables cross-domain data translation but does not address data generation; and other GAN models (e.g., CoGAN~\cite{liu2016coupled}) which can generate multi-domain images simultaneously still require a large amount of data to learn the satisfactory target-domain generator.

\begin{figure}[t]
    \centering
    \includegraphics[width=.85\linewidth]{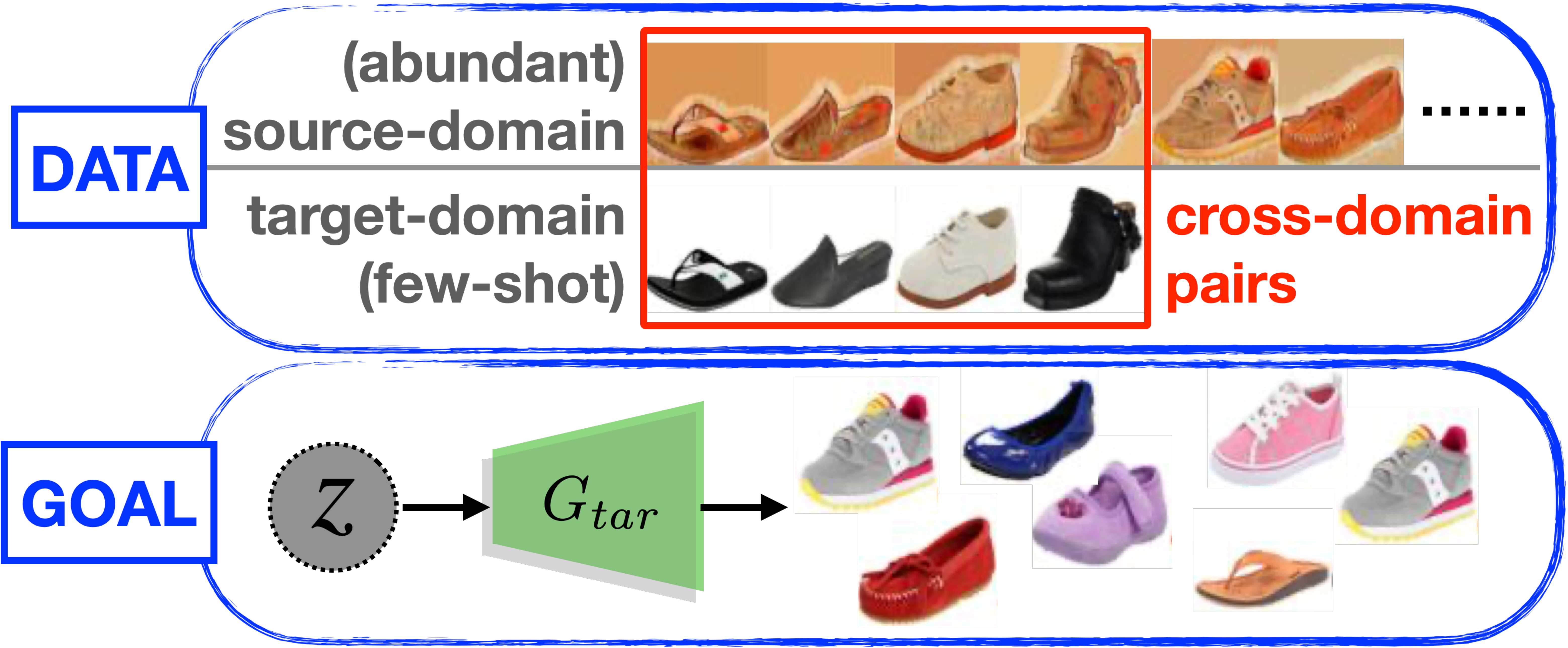}
    \vspace{-1em}
    \caption{Illustration of the paired few-shot (PFS) learning and generation problem setup in this paper.
    There are abundant and diverse training data in the source domain but only few training data in the target domain with known correspondence to their source-domain counterparts. Our PFS-GAN uses these cross-domain data and paired few-shot information to train a target-domain generator.
    }
    \label{Fig_PFS_GAN}
    \vspace{-2em}
\end{figure}

As shown in Figure \ref{Fig_PFS_GAN}, this paper aims to learn a target-domain generator with sufficient source-domain data but only few paired target-domain data.
Specifically, we introduce few-shot learning to help transferring the knowledge of generation from source to target domain. Noting that since most existing few-shot learning works focus on classification and use supervised metrics for learning new classes, they are not directly applicable to GAN.
To integrate the concept of few-shot learning into learning target-domain generator, we consider the \textbf{\textit{paired few-shot setting}} where each target-domain sample has a correspondence in the source, which means these two domains share similar contents but have difference appearances. Under this setting, our proposed \textbf{Paired Few-Shot GAN (PFS-GAN)} could transfer the knowledge across domains by domain adaptation, while learning to generate target-domain samples with high diversity and fidelity.
Particularly, the cross-domain knowledge is learned via disentangled representations composed of domain-invariant content features and domain-specific appearance features. While the target-domain generator learns the appearance feature from the few target-domain data, the rich content information transferred from the source domain can thus help improving the diversity of structure generation in the target domain.

%% file: method.tex
\section{Proposed Method: PFS-GAN}\label{sec:method}

\begin{figure*}
    \centering
    \includegraphics[width=0.9\textwidth]{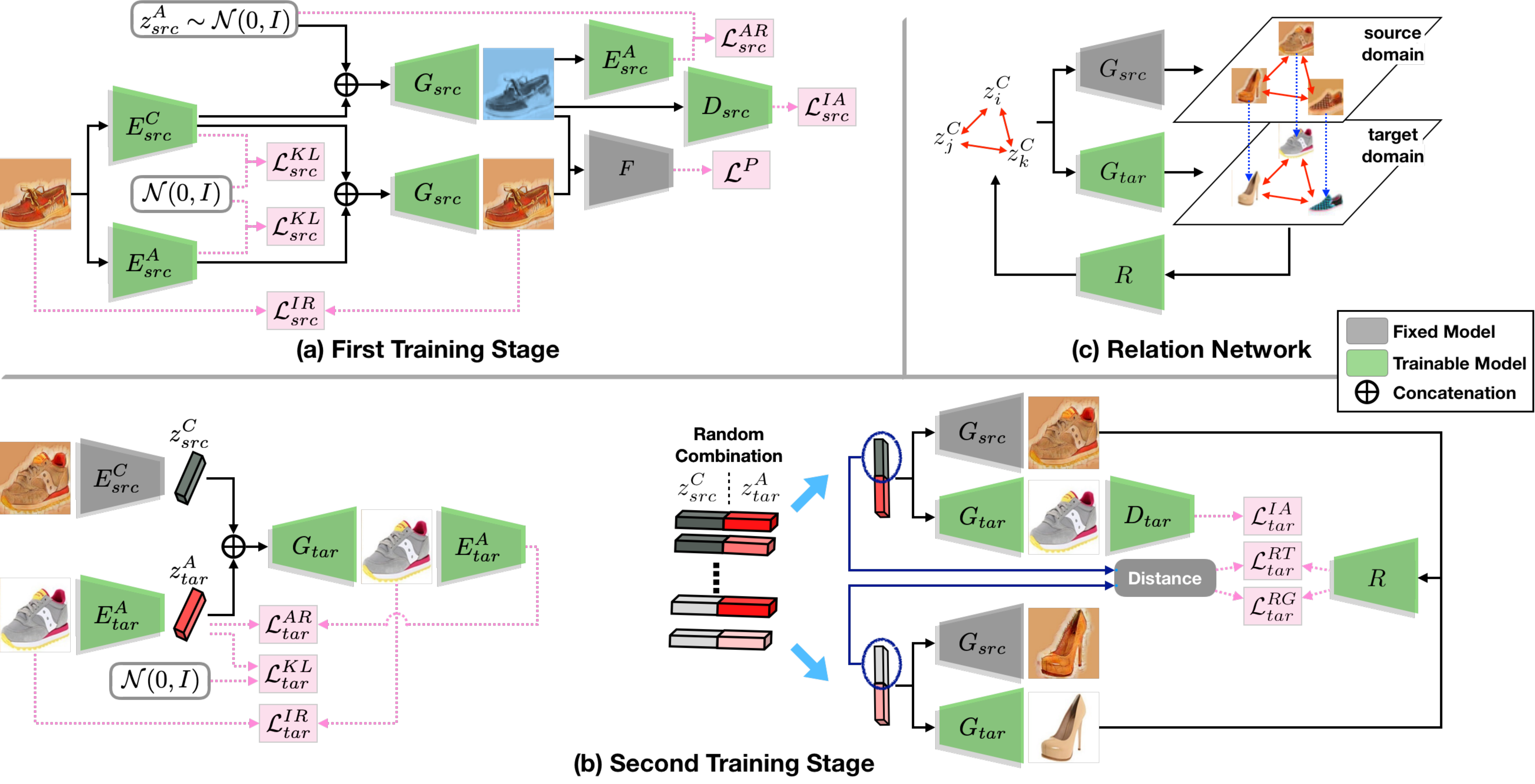}
    \vspace{-1em}
    \caption{Overview of our proposed Paired Few-Shot GAN (\textbf{PFS-GAN}) model. \textbf{(a)} The first training stage: learning source-domain generator and disentanglement (cf. Section~\ref{sec:stage-1}). \textbf{(b)} The second training stage: learning target-domain generator (cf.  Section~\ref{sec:stage-2}). \textbf{(c)} Relation network leverages the content relation between cross-domain samples for regularizing the content diversity between generated target-domain samples. }
    \label{PFS-GAN}
    \vspace{-1em}
\end{figure*}

\vspace{-.9em}
Given a cross-domain dataset built upon the source-domain  $X_{src} = \{ x_{src,i}\}^{N_{src}}_{i=1}$ and target-domain  $X_{tar} = \{ x_{tar,i}\}^{N_{tar}}_{i=1}$ samples where $N_{tar} \ll N_{src}$, our problem scenario assumes that for each $x_{tar, i}$ there is a paired source-domain sample $x_{src, \kappa(i)}$ in which $x_{tar, i}$ and $x_{src, \kappa(i)}$ have the similar content/structure but different appearance/texture, where $\kappa(\cdot)$ is a one-to-one mapping function to obtain the index correspondence across domains. All these cross-domain pairs are denoted as $X^{Pair}=\{x_{tar,i}, x_{src,\kappa(i)}\}^{N_{tar}}_{i=1}$. In order to transfer the richer content from the source to the target domain and make the target-domain generator capable of generating target-domain samples with more diverse content, we first factor out the content information from the source-domain samples (i.e. \textbf{disentanglement}), then tackle the challenges of domain adaptation and few-shot setting to benefit the target-domain generator. Our proposed PFS-GAN consists of generators $\{G_{src}, G_{tar}\}$, discriminators $\{D_{src}, D_{tar}\}$, source content encoder $E^{C}_{src}$, appearance encoders $\{E^{A}_{src}, E^{A}_{tar}\}$ and the relation network $R$. We now detail the two training stages for PFS-GAN (as depicted in Figure~\ref{PFS-GAN}) in the following.

\vspace{-.75em}
\subsection{\textbf{First stage: learning $G_{src}$ and disentanglement}}\label{sec:stage-1}
\vspace{-.25em}
The disentanglement of the source-domain latent space into content and appearance parts is realized by a framework based on VAE~\cite{kingma2013auto} and GAN. Given a source-domain sample $x_{src}$, source-domain content encoder $E^C_{src}$ and appearance encoder $E^A_{src}$ map it into the content feature $z^C_{src}$ and the appearance feature $z^A_{src}$ respectively, where $\{z^C_{src}, z^A_{src}\}$ can be used to reconstruct $x_{src}$ by source-domain generator $G_{src}$ with their distribution modeled by standard normal distribution $\mathcal{N}(0,I)$. The source-domain image reconstruction loss $\mathcal{L}^{IR}_{src}$ and the KL-divergence loss $\mathcal{L}^{KL}_{src}$ are thus defined as:
\begin{equation*}
    \begin{aligned}
        \mathcal{L}^{IR}_{src} =& \mathbb{E}_{x_{src} \sim X_{src}} \left \| G_{src}(z^C_{src}, z^A_{src}) - x_{src}\right \| \\
        \mathcal{L}^{KL}_{src} =& \mathbb{E}_{x_{src} \sim X_{src}} \left [ D_{KL}(E^{C}_{src}(x_{src})|| \mathcal{N}(0,I)) \right ] \\
        &+ \mathbb{E}_{x_{src} \sim X_{src}}\left [ D_{KL}(E^{A}_{src}(x_{src})|| \mathcal{N}(0,I)) \right ]\\
    \end{aligned}
\end{equation*}
For better encouraging the disentanglement between the content and appearance features, two objectives are introduced additionally. First, given two images generated by the same content vector $z^C_{src}$ but different appearance vectors ($z^A_{src}$ and $z^{{A}'}_{src}$), they should have high similarity on the content/structure. Since the higher layers of the ImageNet-pretrained VGG network tend to represent the high-level structure
content of an image~\cite{gatys2016image}, we use the perceptual loss $\mathcal{L}^P$~\cite{johnson2016perceptual} to ensure $z^C_{src}$ maintaining the content information:
\begin{equation*}
    \mathcal{L}^{P} = \left \| F(G(z^C_{src}, z^A_{src}))-F(G(z^C_{src}, z^{{A}'}_{src}))\right \|
\end{equation*}
where $F$ is pretrained VGG network up to \texttt{relu3\_3} layer. Second, for preventing appearance vector $z^A_{src}$ from being ignored by the generator $G_{src}$, we have the latent reconstruction loss $\mathcal{L}^{AR}_{src}$ on the appearance feature $z^A_{src}$:
\begin{equation*}
    \mathcal{L}^{AR}_{src} = \mathbb{E}_{z^A_{src} \sim \mathcal{N}(0,I)} \left\|  E^A_{src}(G_{src}(z^C_{src},z^A_{src})) - z^A_{src}\right\|
\end{equation*}
in which it means that for a synthetic image $G_{src}(z^C_{src}, z^A_{src})$, we should be able to obtain $z^A_{src}$ from it via the appearance encoder $E^A_{src}$. Note that here $z^A_{src} \sim \mathcal{N}(0,I)$ and $z^C_{src}$ can be any content vector extracted from the source-domain samples.
Moreover, we improve the realness of the synthetic $\tilde{X}_{src}$ via adversarial learning (based on the hinge loss~\cite{tran2017hierarchical}), where 
\begin{equation*}
    \begin{aligned}
    \mathcal{L}^{IA,D}_{src} =& \mathbb{E}_{x_{src} \sim X_{src}}[1-D_{src}(x_{src})]_{+}\\ &+
    \mathbb{E}_{\tilde{x}_{src} \sim \tilde{X}_{src}}[1+D_{src}(\tilde{x}_{src})]_{+}\\
    \mathcal{L}^{IA,G}_{src} =& 
    -\mathbb{E}_{\tilde{x}_{src} \sim \tilde{X}_{src}}[D_{src}(\tilde{x}_{src})]
    \end{aligned}
\end{equation*}
are used to update $D_{src}$ (discriminator) and $G_{src}$ respectively.

\vspace{-.75em}
\subsection{Second stage: learning target generator $G_{tar}$}\label{sec:stage-2}
\vspace{-.25em}
Once the source-domain generator $G_{src}$ and its latent space disentanglement are learned, we now aim to train the target-domain generator $G_{tar}$ which combines the diverse content inherited from the source with the target-domain specific appearance to synthesize target-domain samples. To this end, we also require the disentanglement of target-domain latent space but now it is realized by $\{E^C_{src}. E^A_{tar}\}$ and $G_{tar}$ (noting that we use $E^C_{src}$ here as two domains share the same content). First, as a target-domain sample $x_{tar,i}$ shares the same content information with its source-domain correspondence $x_{src, \kappa(i)}$, the image reconstruction loss $\mathcal{L}^{IR}_{tar}$ is defined as:
\begin{equation*}
    \mathcal{L}^{IR}_{tar} = \sum_{i=1}^{N_{tar}} \left \| G_{tar}(E^C_{src}(x_{src,\kappa(i)}), E^A_{tar}(x_{tar,i})) - x_{tar,i}\right \| \\
\end{equation*}
Second, we would also like to regularize the distribution of target-domain appearance features $z^A_{tar}$ by $\mathcal{N}(0,I)$. However, as there are only few-shots for target-domain data which could be problematic to directly apply the regularization on their distribution, we adopt the data augmentation to increase the appearance variation of target-domain data, where the augmentation is performed by randomly shifting the chromatic channels in the Lab color space. The KL-divergence loss $\mathcal{L}^{KL}_{tar}$ is defined on the augmented data samples $X_{tar}^{aug}$:
\begin{equation*}
    \mathcal{L}^{KL}_{tar} = \mathbb{E}_{x_{tar}\sim X_{tar}^{aug}}\left [ D_{KL}(E^{A}_{tar}(x_{tar})|| \mathcal{N}(0,I)) \right ]
\end{equation*}
Third, we also use the latent reconstruction loss $\mathcal{L}^{AR}_{tar}$ to encourage the enhance the disentanglement:
\begin{equation*}
  \mathcal{L}^{AR}_{tar} = \mathbb{E}_{z^A_{tar}\sim E^A_{tar}(X_{tar}^{aug})} \left\|  E^A_{tar}(G_{tar}(z^C_{src},z^A_{tar})) - z^A_{tar}\right\|
\end{equation*}
where now $z^A_{tar}$ is sampled from $E^A_{tar}(X_{tar}^{aug})$ and $z^C_{src}$ can be any content vector extracted from $x_{src} \in X^{Pair}$.
Fourth, we also impose adversarial loss on the synthetic target-domain samples $X^{syn}_{tar}$ which use source content $z^C_{src} \sim E^C_{src}(X_{src})$ and target appearance features $z^A_{tar} \sim E^A_{tar}(X_{tar})$, where
\begin{equation*}
    \begin{aligned}
    \mathcal{L}^{IA,D}_{tar} =& \mathbb{E}_{x_{tar}\sim X_{tar}}[1-D_{tar}(x_{tar})]_{+}\\ &+
    \mathbb{E}_{x_{tar}\sim X^{syn}_{tar}}[1+D_{tar}(x_{tar})]_{+}\\
    \mathcal{L}^{IA,G}_{tar} =& -\mathbb{E}_{x_{tar}\sim X^{syn}_{tar}}[D_{tar}(x_{tar})]\\
    \end{aligned}
\end{equation*}
are used to update $D_{tar}$ (discriminator) and $G_{tar}$ respectively. 

Lastly, in order to better enforce the generated target-domain samples on having the rich content adapted from the source, we particularly propose a novel \textbf{Relation Loss} which leverages the relation among content vectors of cross-domain samples to regularize the content diversity between generated target-domain samples, where the idea is shown in Figure~\ref{PFS-GAN}(c): the relations between different content vectors $z^C$ (i.e. pairwise distance) should be well reflected on the content similarity among the related generated samples, which we would like to have such property in both source and target domains. To this end, we first define the content similarity between cross-domain images: Given any cross-domain image pair $\{x_{src, j}, x_{tar, i}\} \notin X^{Pair}$ where $j\neq \kappa(i)$, their content similarity is computed by
\begin{equation*}
   D_c(x_{src, j}, x_{tar, i}) = \left\| E^C_{src}(x_{src, j}) - E^C_{src}(x_{src, \kappa(i)})\right\|
\end{equation*}
as $x_{tar, i}$ and its paired $x_{src, \kappa(i)}$ should have the same content information. We then train a relation network $R$ learning to project the cross-domain image pair into a value representing their content similarity, which exactly equals to the L2 distance between their corresponding content vectors, with the objective:
\begin{equation*}
    \mathcal{L}^{RT}_{tar} = \sum_{i}^{N_{tar}}\sum_{j\neq\kappa(i)}^{N_{src}} \left \|R(x_{src, j}, x_{tar, i})-D_c(x_{src, j}, x_{tar, i}) \right\|
\end{equation*}
Once the relation network $R$ is learnt, we use it to regularize the content diversity between the generated target-domain samples: Given a cross-domain pair of generated samples, $\{\tilde{x}_{src}=G_{src}(z^C_{src, i}, z^A_{src}), \tilde{x}_{tar}=G_{tar}(z^C_{src, j}, z^A_{tar})\}$ where $i\neq j$, $z^C_{src,i}$ and $z^C_{src,j}$ are taken from the source-domain images by $E^C_{src}$,  $z^A_{src}\sim \mathcal{N}(0,I)$, and $z^A_{tar}\in E^A_{tar}(X_{tar})$, their content similarity computed by $R$ should be equal to the L2 distance between $z^C_{src, i}$ and $z^C_{src, j}$, leading to the relation loss:
\begin{equation*}
    \mathcal{L}^{RG}_{tar} = \mathbb{E} \left\| R(\tilde{x}_{src}, \tilde{x}_{tar}) - D_c(\tilde{x}_{src}, \tilde{x}_{tar}) \right\|
\end{equation*}
where we sample many cross-domain pairs of $\{\tilde{x}_{src},\tilde{x}_{tar}\}$. The gradient of $\mathcal{L}^{RG}_{tar}$ helps update $G_{tar}$ and enforce it to generate target-domain samples with sufficient content diversity.

\begin{table*}[ht]
\caption{Quantitative comparison of target-domain generation among various approaches under different experimental settings and metrics.}
\vspace{-.8em}
\centering
\adjustbox{max width=\textwidth}{%
\begin{tabular}{c|c c|c c|c c|c c|c c|c c|c c|c c}

\cline{1-17}
    \begin{tabular}[c]{@{}c@{}}Experiment \end{tabular}
    & \multicolumn{4}{c|}{Edge2Shoes (E2S)}& \multicolumn{4}{c|}{Style2Shoes (S2S)}
    & \multicolumn{4}{c|}{Face2Face (F2F)}& \multicolumn{4}{c}{Sketch2Face (S2F)}\\ 
    \begin{tabular}[c]{@{}c@{}} \ Synthesis Manner  \end{tabular}
    & \multicolumn{2}{c}{Rand} & \multicolumn{2}{c|}{Syn} & \multicolumn{2}{c}{Rand}
    & \multicolumn{2}{c|}{Syn}  & \multicolumn{2}{c}{Rand} & \multicolumn{2}{c|}{Syn}
    & \multicolumn{2}{c}{Rand} & \multicolumn{2}{c}{Syn}  \\ \cline{1-17}
    Metric & FID & KID & FID & KID& FID & KID& FID & KID &  FID & KID&  FID & KID&  FID & KID&  FID & KID\\ \cline{1-17}
    
    BaselineS & 185.6 & 0.37 & - & - & 219.9 & 0.18 & - & - & 143.8 & 0.19 & - & - & 122.6 & 0.20 & - & -\\ 
    BaselineT & 165.0 & 0.29 & - & - & 149.9 & 0.14 & - & - & 111.4 & 0.22 & - & - & 125.0 & 0.19 & - & -\\ 
    
    CoGAN & 197.2 & 0.38 & - & - & 179.0 & 0.14 & - & - & 220.2 & 0.15 & - & - & 195.7 & 0.35 & - & - \\
    UNIT & 135.8 & 0.34 & - & - & 143.3 & 0.27 & - & - & 103.1 & 0.42 & - & - & 93.04 & 0.44 & - & -\\ \cline{1-17}
    
    PFS-GAN$^\ddagger$ & 125.9 & 0.21 & 103.5 & 0.12 & 116.1 & 0.13 & 93.37 & 0.17 & 55.47 & 0.08 & 32.67 & 0.04 & 48.93 & 0.08 & 44.83 & 0.06\\ 
    PFS-GAN$^\dagger$ & \textbf{91.42} & 0.19 & 88.93 & 0.08 & 96.08 & 0.11 & 78.83 & 0.12 & 50.17 & 0.06 & 29.45 & 0.03 & 45.61 & 0.06 & 40.78 & 0.06\\ 
    PFS-GAN & 92.55 & \textbf{0.17} & \textbf{83.31} & \textbf{0.07} & \textbf{79.94} & \textbf{0.08} & \textbf{67.22} & \textbf{0.11} & \textbf{40.76} & \textbf{0.04} & \textbf{25.11} & \textbf{0.02} & \textbf{38.50} & \textbf{0.04} & \textbf{36.95} & \textbf{0.04} \\ \hline
\end{tabular}
}
\vspace{-.5em}
\label{table:Dataset}
\end{table*}

\begin{table*}[ht]
  \centering
  \caption{Qualitative examples produced by various baselines and our proposed PFS-GAN model under different experimental settings.}
  \vspace{-1em}
  \label{table2}
  \begin{tabular}{  c | c | c | c | c}
    \begin{tabular}[c]{@{}c@{}} \end{tabular} & Style2Shoes (S2S) & Edge2Shoes (E2S) & Face2Face (F2F) & Sketch2Face (S2F) \\ \hline
    
    Source Data & 
    \begin{minipage}{.2\textwidth}
      \includegraphics[width=1.\linewidth]{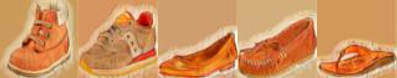}
    \end{minipage} & 
    \begin{minipage}{.2\textwidth}
      \includegraphics[width=1.\linewidth]{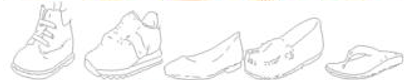}
    \end{minipage} &
    \begin{minipage}{.2\textwidth}
      \includegraphics[width=1.\linewidth]{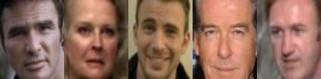}
    \end{minipage} & 
    \begin{minipage}{.2\textwidth}
      \includegraphics[width=1.\linewidth]{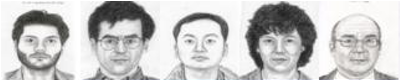}
    \end{minipage} \\ 

    Target Data & 
    \begin{minipage}{.2\textwidth}
      \includegraphics[width=1.\linewidth]{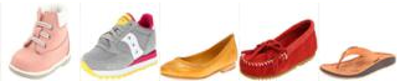}
    \end{minipage} & 
    \begin{minipage}{.2\textwidth}
      \includegraphics[width=1.\linewidth]{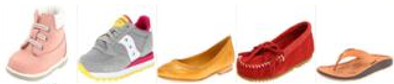}
    \end{minipage} & 
    \begin{minipage}{.2\textwidth}
      \includegraphics[width=1.\linewidth]{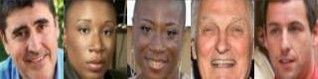}
    \end{minipage} & 
    \begin{minipage}{.2\textwidth}
      \includegraphics[width=1.\linewidth]{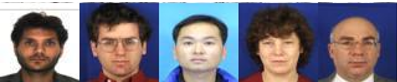}
    \end{minipage} \\ \hline \hline
    
    BaselineS & 
    \begin{minipage}{.2\textwidth}
      \includegraphics[width=1.\linewidth]{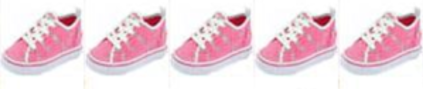}
    \end{minipage} & 
    \begin{minipage}{.2\textwidth}
      \includegraphics[width=1.\linewidth]{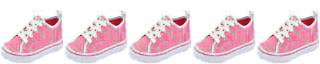}
    \end{minipage} &
    \begin{minipage}{.2\textwidth}
      \includegraphics[width=1.\linewidth]{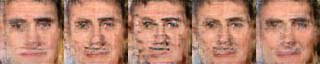}
    \end{minipage} & 
    \begin{minipage}{.2\textwidth}
      \includegraphics[width=1.\linewidth]{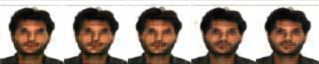}
    \end{minipage} \\ 
   
    BaselineT & 
    \begin{minipage}{.2\textwidth}
      \includegraphics[width=1.\linewidth]{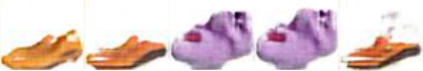}
    \end{minipage} & 
    \begin{minipage}{.2\textwidth}
      \includegraphics[width=1.\linewidth]{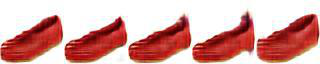}
    \end{minipage} &
    \begin{minipage}{.2\textwidth}
      \includegraphics[width=1.\linewidth]{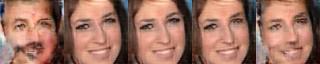}
    \end{minipage} & 
    \begin{minipage}{.2\textwidth}
      \includegraphics[width=1.\linewidth]{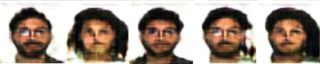}
    \end{minipage} \\ 
    
    CoGAN & 
    \begin{minipage}{.2\textwidth}
      \includegraphics[width=1.\linewidth]{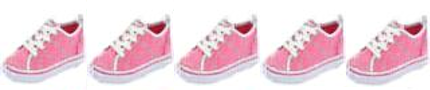}
    \end{minipage} & 
    \begin{minipage}{.2\textwidth}
      \includegraphics[width=1.\linewidth]{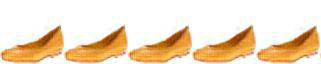}
    \end{minipage} &
    \begin{minipage}{.2\textwidth}
      \includegraphics[width=1.\linewidth]{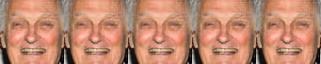}
    \end{minipage} & 
    \begin{minipage}{.2\textwidth}
      \includegraphics[width=1.\linewidth]{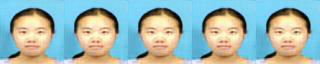}
    \end{minipage} \\ 
   
    UNIT & 
    \begin{minipage}{.2\textwidth}
      \includegraphics[width=1.\linewidth]{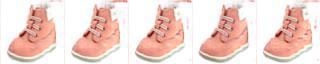}
    \end{minipage} & 
    \begin{minipage}{.2\textwidth}
      \includegraphics[width=1.\linewidth]{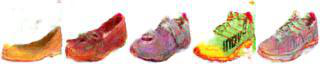}
    \end{minipage} &
    \begin{minipage}{.2\textwidth}
      \includegraphics[width=1.\linewidth]{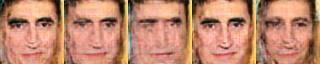}
    \end{minipage} & 
    \begin{minipage}{.2\textwidth}
      \includegraphics[width=1.\linewidth]{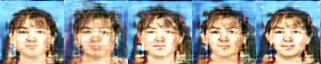}
    \end{minipage} \\ \hline
   
    PFS-GAN & 
    \begin{minipage}{.2\textwidth}
      \includegraphics[width=\linewidth]{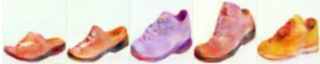}
    \end{minipage} & 
    \begin{minipage}{.2\textwidth}
      \includegraphics[width=\linewidth]{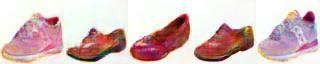}
    \end{minipage} &
    \begin{minipage}{.2\textwidth}
      \includegraphics[width=\linewidth]{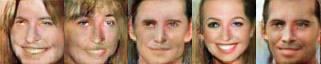}
    \end{minipage} & 
    \begin{minipage}{.2\textwidth}
      \includegraphics[width=\linewidth]{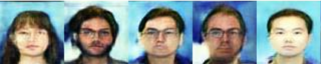}
    \end{minipage} \\  
    
  \end{tabular}
  \vspace{-1.5em}
  \end{table*}

%% file: exp.tex
\vspace{-1em}
\section{Experiments}\label{sec:experiments}
\vspace{-.5em}

\noindent \textbf{\textcolor{blue}{Datasets}}~
Three datasets, \textbf{CUFS}~\cite{wang2008face}, \textbf{UT Zappos50K}~\cite{semjitter}, and \textbf{Facescrub}~\cite{ng2014data}, are used to build four cross-domain experimental settings. UT Zappos50K is a large dataset of shoes images with sketches provided for each real shoes image; Facescrub dataset is composed of face images of 530 celebrities, taken under various poses/angles; 
CUFS dataset consists of image pairs of face portraits and their corresponding sketches. 
Four experimental settings are listed as follows: \textbf{(1) Edge2Shoes} picks 10 real shoes images as the target-domain, while their corresponding sketches with many others (in total 10,000 sketches) are taken as the source-domain; \textbf{(2) Style2Shoes} is almost identical to Edge2Shoes but the sketches are stylized by~\cite{huang2017arbitrary}; \textbf{(3) Face2Face} is based on Facescrub, where the source-domain is composed of photos of 525 celebrities while the target-domain has 10 photos from each of the left 5 celebrities (i.e. adapting diverse poses from source to target); \textbf{(4) Sketch2Face} randomly selects 10 real portraits as target-domain while having all 1,194 sketches as source.

\noindent \textbf{\textcolor{blue}{Baselines}}~Four baselines are used for comparison with our proposed PFS-GAN: \textbf{(1) BaselineS} is a GAN trained from scratch based on target-domain samples only; \textbf{(2) BaselineT} is similar to BaselineS but uses the trained $G_{src}$ as its initialization; \textbf{(3) CoGAN}~\cite{isola2017image} consists of two tuple of GAN trained by the cross-domain data with the weights of their networks partially shared; \textbf{(4) UNIT}~\cite{zhu2017unpaired} is the extension of CoGAN which is also trained by the cross-domain data, with having additional loss to regularize the paired cross-domain images being mapped into the same latent vector by the encoders.

Fr\'{e}chet Inception distance (FID~\cite{heusel2017gans}) and Kernel Inception Distance (KID~\cite{binkowski2018demystifying}) are used (both the lower the better) for quantitative evaluation on the target-domain generators. Two PFS-GAN variants are also used to perform the ablation study: (1) \textbf{PFS-GAN$^{ \dagger}$} removes the relation loss and (2) \textbf{PFS-GAN$^{ \ddagger}$} further removes the adversarial loss in the second training stage. Two manners are used for PFS-GAN to generate target-domain samples: \textbf{Rand} takes both $z^C$ and $z^A$ sampled from $\mathcal{N}(0,I)$ while \textbf{Syn} takes $\{z^C, z^A\}$ sampled from $\{E^C_{src}(X_{src}),E^A_{tar}(X_{tar})\}$ respectively. 
Quantitative results are shown in Table~\ref{table:Dataset}, where our full PFS-GAN clearly has superior performance than the baselines, and the comparison to the variants verifies the contributions of our designed objectives (particularly the relation loss). Moreover, the difference between PFS-GAN$^{ \ddagger}$ and the baselines shows the importance of disentanglement. 
BaselineS and BaselineT perform around the second worst, since they are trained only with few-shot target-domain samples hence easily leading to mode collapse; CoGAN mostly performs the worst as the number of samples for both domains is quite unbalanced thus leading to unstable discriminators; UNIT improves over CoGAN but still can not produce satisfactory results. 
We also have experimented on increasing target-domain samples from 10 to 20 then 100. Take Sketch2Face as an example, PFS-GAN boosts from 38.5 to 33.6 then achieves 28.4 in FID, while other baselines are also improved but still worse than ours by a margin with the issues of unstable training and mode collapse remained.

Table~\ref{table2} provides the qualitative examples. We observe that: BaselineS, BaselineT, and CoGAN generate realistic images, but clearly attempts to memorize the training samples and lose the content diversity (i.e. mode collapse caused by few-shot data). UNIT also suffers from mode collapse (e.g. Style2Shoes and Face2Face), while occasionally produces results with poor quality as in the Sketch2Face case. 
Our PFS-GAN provides favorable results on both content diversity and fidelity, and is able to synthesize images with having the content that is never seen in the target-domain training samples.

%% file: conclusion.tex
\vspace{-1em}
\section{Conclusion}\label{sec:conclusion}
\vspace{-1em}
We propose PFS-GAN to tackle the generative model learning with cross-domain data, where the target-domain has only few-shots provided with paired source-domain samples. PFS-GAN combines the learning disentanglement (domain-invariant content and domain-specific appearance features), domain adaptation, and the cross-domain relation built upon the properties of training data. The target-domain generator from PFS-GAN experimentally shows its capacity on improving the content diversity of generated images and providing superior performance in comparison to several baselines.